\documentclass[journal]{IEEEtran}
\ifCLASSINFOpdf
\else
\fi
\usepackage{graphicx}
\usepackage{url}
\usepackage{amsmath,amssymb}
\usepackage{stfloats}
\usepackage{algorithm}
\usepackage{algorithmicx}
\usepackage{algpseudocode}
\usepackage{mathrsfs}
\usepackage{booktabs}
\usepackage{multirow}
\usepackage{tabu}
\usepackage{xcolor}

\begin{document}
\title{Dynamic Sampling Networks for Efficient Action Recognition in Videos}

\author{Yin-Dong Zheng,
        Zhaoyang~Liu,
        Tong Lu~\IEEEmembership{Member,~IEEE},
        Limin~Wang~\IEEEmembership{Member,~IEEE}
        % <-this % stops a space

\thanks{Y. Zheng, Z. Liu, T. Lu, and L. Wang are with the State Key Laboratory for Novel Software Technology, Nanjing University, Nanjing, 210023, China.}
\thanks{Y. Zheng and Z. Liu equally contribute to this work.}
}

\maketitle

\begin{abstract}
The existing action recognition methods are mainly based on clip-level classifiers such as two-stream CNNs or 3D CNNs, which are trained from the randomly selected clips and applied to densely sampled clips during testing.
However, this standard setting might be suboptimal for training classifiers and also requires huge computational overhead when deployed in practice.
To address these issues, we propose a new framework for action recognition in videos, called {\em Dynamic Sampling Networks} (DSN), by designing a dynamic sampling module to improve the discriminative power of learned clip-level classifiers and as well increase the inference efficiency during testing.
Specifically, DSN is composed of a sampling module and a classification module, whose objective is to learn a sampling policy to on-the-fly select which clips to keep and train a clip-level classifier to perform action recognition based on these selected clips, respectively.
In particular, given an input video, we train an observation network in an associative reinforcement learning setting to maximize the rewards of the selected clips with a correct prediction.
We perform extensive experiments to study different aspects of the DSN framework on four action recognition datasets: UCF101, HMDB51, THUMOS14, and ActivityNet v1.3.
The experimental results demonstrate that DSN is able to greatly improve the inference efficiency by only using less than half of the clips, which can still obtain a slightly better or comparable recognition accuracy to the state-of-the-art approaches.
\end{abstract}


% \fi
\IEEEpeerreviewmaketitle

% ############################################ Introduction
\section{Introduction}
\label{intr}

\IEEEPARstart{A}{ction} recognition in videos has drawn enormous attention from the computer vision community in the past decades~\cite{WangS13a,SimonyanZ14,KarpathyTSLSF14,TranBFTP15,TSN-J,CarreiraZ17,R2+1D,ArtNet,s3d,nonlocal}, partially due to its applications in many areas such as video surveillance, content-based analysis, and human-computer interaction.
Recently, deep learning methods have witnessed great success for many image-based tasks, such as image classification~\cite{HeZRS16,KrizhevskySH12}, object detection~\cite{GirshickDDM14}, semantic segmentation~\cite{LongSD15}, and pose estimation~\cite{HeGDG17}.
These deep models have also been introduced into the video domain for action recognition~\cite{TSN-J, CarreiraZ17}, which has shown superior performance to the traditional hand-crafted representations~\cite{WangS13a}.
However, unlike the static image, the extra temporal dimension of the video increases the difficulty of developing an effective deep action recognition approach.
When proposing an action recognition solution,  it is necessary to carefully consider the inherent properties of temporal information.

\begin{figure}
\centering
\includegraphics[width=8.3cm]{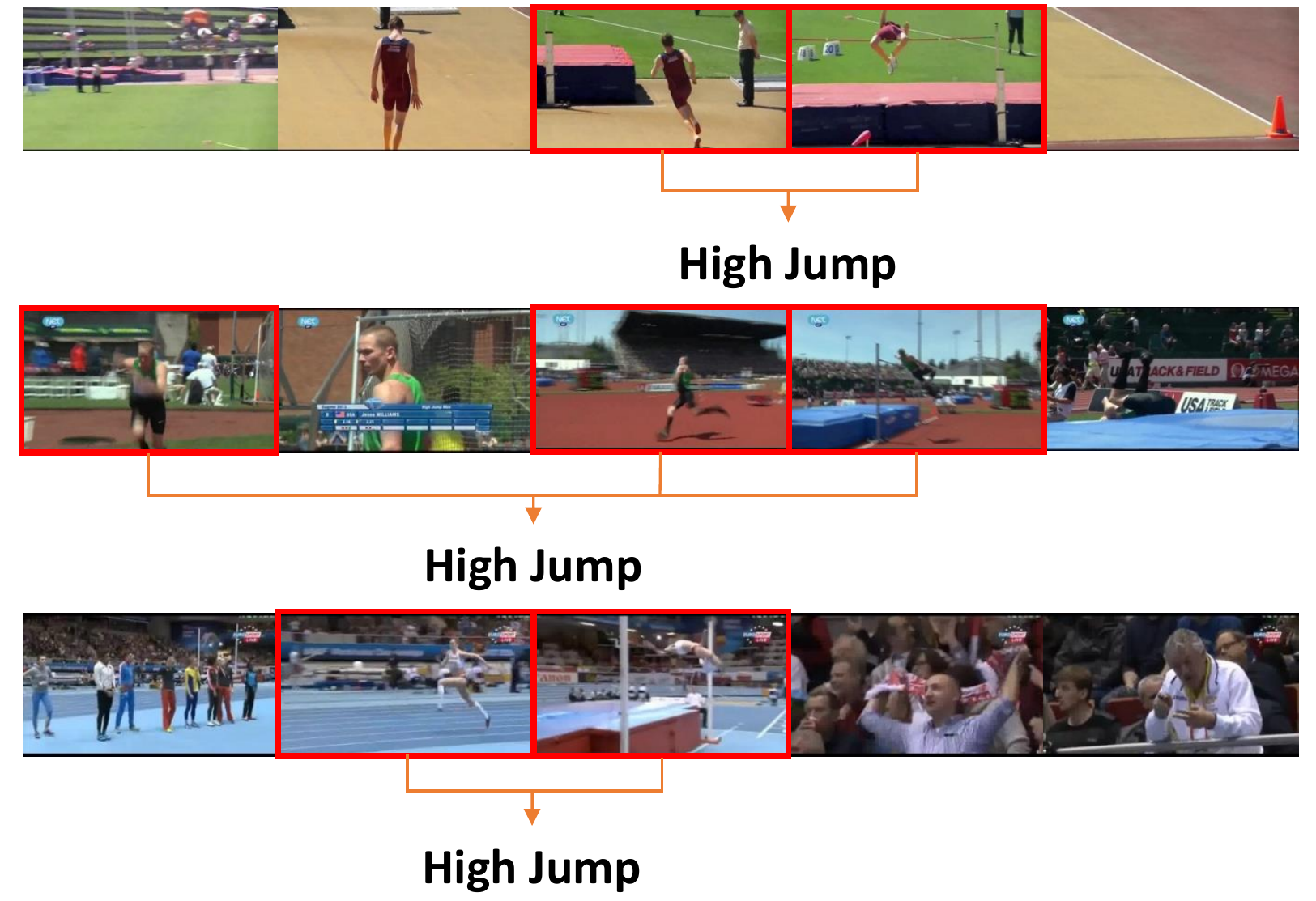}
\caption{\textbf{Dynamic clip sampling for action recognition.} DSN framework dynamically selects a subset of clips from video for action recognition.
It not only improves the accuracy of the final recognition but also greatly improves the inference efficiency during the testing phase.}
\label{fig:intr}
\end{figure}

Currently, deep learning methods for action recognition are mostly based on short-term classifiers, due to modeling complexity and computational overhead of directly learning a long-term video-level classifier. These short-term classifiers are with architectures of 2D CNNs~\cite{lecun-98} or 3D CNNs~\cite{JiXYY13}, which operate on a single frame (e.g., two-stream CNNs~\cite{SimonyanZ14}) or a short clip of multiple consecutive frames (e.g. C3D~\cite{TranBFTP15}, I3D~\cite{CarreiraZ17}, ARTNet~\cite{ArtNet}).
In the training phase, these short-term classifiers are optimized on a single randomly sampled clip~\cite{TranBFTP15,SimonyanZ14} or a set of uniformly sampled clips from the whole video~\cite{TSN-J}.
In the test phase, to gather the information from the whole video, these trained short-term classifiers are applied in a fully convolutional manner on densely sampled video clips. The recognition results from densely testing are simply averaged to yield the final recognition results.
However, there are two critical issues in the above framework when training short-term classifiers  {\em randomly} and testing these classifiers  {\em densely}.

First, training these short-term classifiers from randomly sampled clips may not be the best approach from the training perspective.
As depicted in Fig.~\ref{fig:intr}, the distribution of action-related information is uneven across the video.
In this sense, some clips in a video may contribute more to action recognition than others.
Furthermore, the distribution of clip-level importance may vary among different input videos.
Therefore, given an input video, we need to on-the-fly select the most important clips to recognize the action in the video.
This dynamic and adaptive sampling module will force short-term classifiers to focus on these more discriminative clips which are also beneficial to train more powerful classifiers.

Second, from the testing perspective, it is expensive to apply these trained short-term classifiers in a fully convolutional manner on densely sampled clips.
Meanwhile, simply averaging from these dense prediction scores may also be problematic because these recognition results from less relevant clips will over-smooth the final recognition result.
It is well known the clips in the video are redundant and semantics varies slowly along the temporal dimension.
Therefore, it is possible to test these short-term classifiers only on a subset of densely sampled clips, which can greatly improve inference efficiency and possibly improve the final recognition accuracy.

To mitigate the above issues, we propose a new principled framework called as {\em Dynamic Sampling Network} (DSN) for action recognition in videos based on a section based selection policy.
The primary objective is to train an optimized sampling strategy that is able to maintain the accuracy of action recognition while select very few clips based on each input video.
Specifically, the DSN framework is composed of a {\em sampling module} and a {\em classification module}, which are optimized in an alternating manner.
The sampling module is composed of an observation network and a policy maker.
The observation network takes several clips as inputs and generates a posterior probability, and the policy maker determines which clips to select based on the posterior probability distribution.
The selected clips will be fed into the classification module, and the rewards based on the predictions will guide the training of the observation network.
In addition, based on the sampling module output, the classification module will be fine-tuned on the video dataset, by focusing more on these clips with high probabilities.
In implementation, to improve inference efficiency, the complexity of the sampling module is much smaller than that of the classification module.
When a deploying DSN in practice, the trained sampling module will dynamically determine a small subset of clips to be fed into the classification module, thus greatly improving the efficiency of the whole action recognition system.

Concerning the training of our DSN, we formulate the framework of DSN within an associative reinforcement learning~\cite{associative} where all the decisions are made in a single step given the context.
Associative reinforcement learning is different from full reinforcement learning~\cite{SuttonB98}.
Full reinforcement learning is a multi-step Markov decision process.
Through multi-step exploration, agent continuously observes the environment and makes decisions based on state history, and gets rewards.
Associative reinforcement learning is also called $N$-armed contextual bandit~\cite{bandit} which is a simplified reinforcement learning algorithm derived from the bandit problem.
In $N$-armed contextual bandit setting, the agent selects one from $N$ levers by observing the environment and gets a reward.
When the environment changes, the reward of each lever also changes.
Since a full reinforcement learning requires complex reward design and is prone to over-fitting with a single label, it is difficult to train a multi-step Markov decision process solely with single label supervision.
We reduce our designed sampling policy to a single-step Markov decision process, which could be easily optimized with an associative reinforcement learning.
The training objective of the sampling module is to maximize the reward for the selected clip, where the reward is calculated based on the action classifier output and groundtruth.

In experiment, we use ResNet2D-18 as the observation network and use ResNet3D-18 or R(2+1)D-34 as the short-term action classifier to implement DSN framework.
We test DSN on the datasets of UCF101~\cite{abs-1212-0402}, HMDB51~\cite{KuehneJGPS11}, THUMOS14~\cite{THUMOS14} and ActivityNet v1.3~\cite{anet}.
Since the dataset is small, DSN has been pre-trained on Kinetics.
DSN obtains a performance improvement over previous state-of-the-art approaches.
More importantly, DSN applies the short-term classifier only on a smaller number of frames, which greatly reduces the computational overhead while still yields slightly better recognition accuracy than previous approaches.

\begin{figure*}[ht]
  \centering
  \includegraphics[width=1\textwidth]{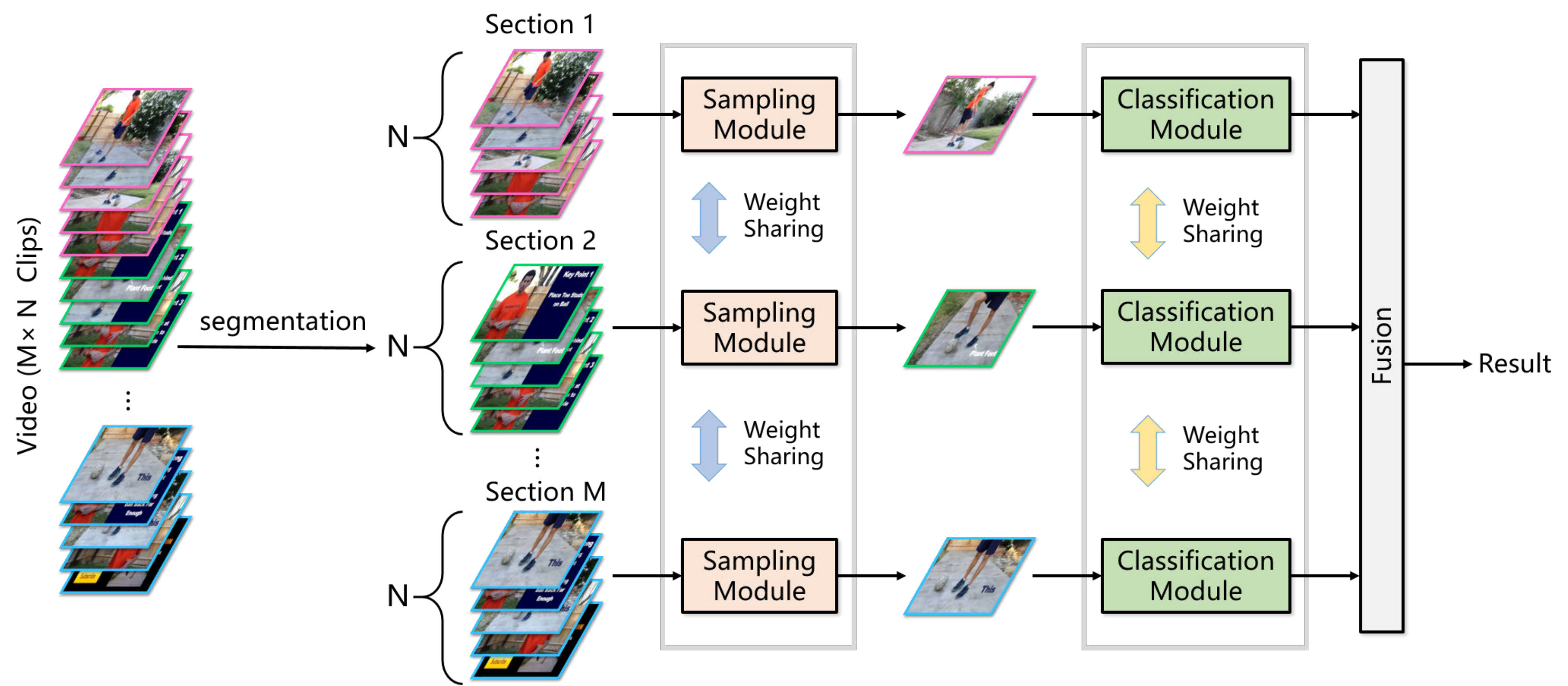}
  \caption{\textbf{Dynamic Sampling Network (DSN).} We devise a dynamic clip sampling strategy, termed as {\em section based selection}, to build our DSN framework. The DSN first divides each video into several sections of equal duration, and then performs dynamic sampling in each section for efficient video recognition. The sampling module and classification module share weights across all sections. Details on the design of sampling module and classification module could be found in Figure~\ref{fig:arch}. }
  \label{fig:arch2}
\end{figure*}

% ################################################### Related works
\section{Related Work}
\label{sec:rw}

\subsection{Deep Learning in Action Recognition}
The breakthrough by AlexNet~\cite{KrizhevskySH12} in image classification started the booming of deep learning in computer vision.
Researchers were inspired to exploit deep network architectures for action recognition in videos~\cite{JiXYY13,KarpathyTSLSF14,SimonyanZ14,Wang0T15,CarreiraZ17,HuangRMTPFN18,TranBFTP15,R2+1D,ArtNet,LiuLWWTWLHL20,Feichtenhofer0M19}.
Ji \emph{et al.}~\cite{JiXYY13} extended 2D CNNs to 3D domain for action recognition and tested the proposed models on small datasets.
To support a huge amount of data needed for training deep networks, Karpathy \emph{et al.} collected a large-scale dataset (Sports-1M)~\cite{KarpathyTSLSF14} with weak tag label from Youtube.
Simonyan \emph{et al.}~\cite{ SimonyanZ14} proposed a two-stream network of spatial and temporal networks that capture appearance and motion information, respectively.
Carreira \emph{et al.}~\cite{CarreiraZ17} proposed I3D by inflating the 2D filters in 2D CNNs into spatiotemporal kernels.
Huang \emph{et al.}~\cite{ HuangRMTPFN18} studied the effect of short-term temporal information on action recognition.
Tran \emph{et al.}~\cite{TranBFTP15} studied 3D CNNs on realistic and large-scale video datasets. Then they further decomposed the 3D convolutional filters into separate spatial and temporal components to obtain R(2+1)D~\cite{R2+1D}, which can improve recognition accuracy.

Meanwhile, several works~\cite{NgHVVMT15,DonahueHGRVDS15,ZhuHSCQ16,TSN-J,WangXLG17,FanXZYGY18} attempted to utilize long-term temporal information for action recognition.
Ng \emph{et al.}~\cite{NgHVVMT15} and Donahue \emph{et al.}~\cite{DonahueHGRVDS15} combined information across the whole video over longer time periods by recurrent neural networks (LSTM).
Zhu \emph{et al.}~\cite{ZhuHSCQ16}  proposed KVMF to filter out irrelevant information by identifying key volumes in the sequences.
Wang \emph{et al.}~\cite{TSN-J} designed a temporal segment network (TSN) that used sparsely sampled frames to aggregate the temporal information of video. They also developed an UntrimmedNet~\cite{WangXLG17} which utilized the attention model to aggregate information for learning from the untrimmed datasets.

\subsection{Efficient Action Recognition}
Neural networks are typically cumbersome, and there is significant redundancy for deep learning models.
To run deep neural network models on mobile devices, we must effectively reduce the storage and computational overhead of the networks.
Parameter pruning and sharing~\cite{SrinivasB15,HanPTD15,ChenWTWC15,LebedevL16,WuNKRDGF18} reduced redundant parameters which are not sensitive to the performance.
Knowledge distillation~\cite{HintonVD15,RomeroBKCGB14,BalanRMW15,LuoZLWT16,ChenGS15} trained a compact neural network with distilled knowledge of a large model.

Efficient network architectures proposed recently such as MobileNet~\cite{HowardZCKWWAA17} and $\epsilon$-ResNet~\cite{YuYR18} were developed for training compact networks.
Wu \emph{et al.}~\cite{WuNKRDGF18} used REINFORCE to learn optimal block dropping policies in ResNet.
ECO~\cite{ZolfaghariSB18} greatly improved recognition speed by manually designing the stacking of sample frames.
SlowFast~\cite{Feichtenhofer0M19} obtained spatial semantics by constructing a slow pathway and obtains action semantics through a very lightweight Fast pathway, which greatly reduces computational overhead compared to traditional two-stream networks.
Yu \emph{et al.}~\cite{YuKC10} utilized local appearance and structural information to achieve real-time action recognition.
Korbar \emph{et al.}~\cite{SCSampler} constructed an oracle sampler and trains a clip sampler with performance close to the oracle sampler by using knowledge distillation. Although their idea is similar to ours, they train sampler and classifier in a separate way.

These methods mainly focus on reducing the complexity of the network architecture or reducing the overhead of inference through some artificially designed methods to improve the speed of inference of the network.
Our work provides a novel perspective for reducing the computational overhead of inference by training a lightweight clip sampler within a new reinforcement learning framework. This new framework allows us to jointly train the clip sampler and classifier.

\subsection{Reinforcement Learning in Videos}
Deep reinforcement learning has made many dramatic advances in the field of video understanding.
Gao \emph{et al.}~\cite{GaoYN17a} used reinforcement learning to train an encoder-decoder network for action anticipation.
Yeung \emph{et al.}~\cite{YeungRMF16} utilized reinforcement learning for action detection.
Fan \emph{et al.}~\cite{ FanXZYGY18} proposed an end-to-end reinforcement approach that they developed an agent to decide which frame to watch at the next time step or to stop watching the video to make classification decision at the current time step.
Wang \emph{et al.}~\cite{WangCWWW18} proposed a novel hierarchical reinforcement learning framework for video captioning. Wu \emph{et al.}~\cite{AdaFrame} proposed the AdaFrame framework to determine which frames based on the LSTM~\cite{lstm} module, and the whole framework was trained by policy gradient methods.
Dong \emph{et al.}~\cite{DongZT19} used BiLSTM and reinforcement learning to select clips to improve performance.

Similar to the these approaches, our DSN also uses reinforcement learning to train our selection module for efficient action recognition.
Notably different from them, we carefully design a section based selection policy to leverage the uniform prior to regularize the selection process. In addition, our designed selection policy will reduce to a single-step Markov decision process, which greatly relieves the training difficulty of reinforcement learning algorithm merely with a single label.
Attention-Aware Sampling~\cite{DongZT19} proposed an agent algorithm to select discriminative clips based on the clip-level features. It only improved performance and failed to save computational overhead as it passed all clips into networks for feature extraction.
However, due to the sampling module is lightweight, the computational overhead of the sampling module is very small compared with the classification module. So, our DSN can greatly reduce computational overhead while still be able to maintain or even improve performance.

\begin{figure*}[t]
  \centering
  \includegraphics[width=1\textwidth]{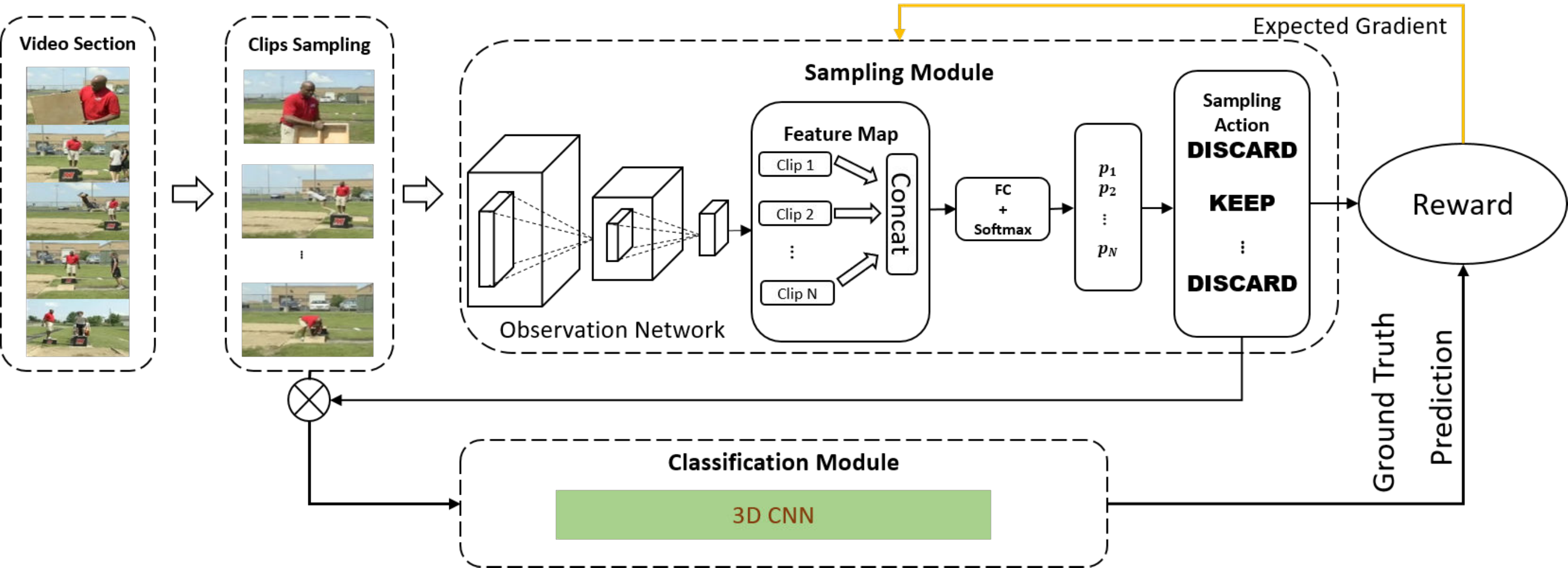}
  \caption{\textbf{Sampling module and classification module in each section.} First, $N$ clips are uniformly sampled from each section. Then these clips are fed into the sampling module separately to detect and rank the important clips. The sampling module evaluates the importance of each clip and decides to keep or discard each clip. After that, the kept clips are fed into the classification module. $\bigotimes$ indicates that clips are discarded or kept according to the sampling action. Predictions from all sections are fused to produce the final prediction. The dynamic sampling network is designed to yield the final recognition results with better performance by keeping as few clips as possible.}
  \label{fig:arch}
\end{figure*}

\section{Dynamic Sampling Networks}
\label{sec:dsn}

In this section, we give a detailed description on our proposed {\em Dynamic Sampling Networks} (DSN).
First, we provide a general overview of our DSN framework.
Then, we present the technical details of the sampling and classification module.

\subsection{Overview of DSN}
As shown in Fig.~\ref{fig:arch2}, given a video $V$, we first divide the video into $M$ sections $\{S^1, S^2, \cdots, S^M\}$ of equal duration without overlap and then uniformly sample $N$ short clips  $C^m=\{c^m_1,c^m_2,...,c^m_N\}$ in section $S^m$.
We sample a total of $M \times N$ clips from the whole video.
The DSN framework is composed of a {\em sampling module} and a {\em classification module}, which performs clip selection in each section and recognize the action class of selected clips, respectively.

Specifically, for the sampling module, it takes the sampled clips from each section as inputs and selects a single clip in each section independently:
\begin{equation}
G^m =\mathcal{G}(c_1^m, c_2^m, \cdots, c_N^m; \Theta),
\label{equ:g}
\end{equation}
where $G^m$ is the selected clip in section $S_m$ by the sampling module $\mathcal{G}$, and $\Theta$ is the model parameter of the sampling module.
Then the output of the sampling module will be fed into the classification module to perform action recognition:
\begin{equation}
H = \mathcal{H}(G^1,\cdots,G^M; \Phi),
\label{equ:h}
\end{equation}
where $H$ is the output of classification module $\mathcal{H}$, representing the prediction result of the video $V$, and $\Phi$ is the model parameter of classification module.
The sampling module dynamically selects a subset of clips for each video and enables the classification module to concentrate on these most discriminative clips. It can greatly improve the inference efficiency when deploying the trained models in practice.
We will present the technical details of modules $\mathcal{G}$ and $\mathcal{H}$, and then describe the training process for optimizing model parameters $\Theta$ and $\Phi$.

{\bf Discussion.} Our DSN presents a dynamic clip sampling policy, called as {\em section based selection}, for efficient video recognition. This section based clip selection is mainly inspired by the work of TSN~\cite{TSN-J}, and our DSN could be viewed as a dynamic version of TSN. This dynamic clip sampling exhibits several important properties: {\em First}, similar to the TSN framework, our DSN leverages a uniform prior to our selection process. We argue that an action instance could be divided into different courses, and each course contains its own important and discriminative information for action recognition. Therefore, our section based selection aims to regularize the selected clips to cover the entire duration of the whole video. {\em Second}, we perform clip selection for each section independently in our designed sampling policy. We analyze that clips of different sections describe different courses of an action instance, and the selection result in a section would have little effect on the other sections. In this sense, our sampling module only needs to focus on distinguishing similar clips in a local section, which would reduce the complexity of the sampling module and increase its flexibility for easy deployment. {\em Finally}, our designed sampling policy will reduce to a single-step Markov decision process, which could be easily optimized with an associative reinforcement learning algorithm as described in Section~\ref{sec:optimization}. Instead, without our section based selection, it is difficult to train a multi-step Markov decision process solely with single label supervision, as a normal reinforcement learning requires complex reward design and is also prone to over-fitting with training data.
%-------------------------------------------------------------------------
\subsection{Sampling Module}
\label{sec:sampling}

The sampling module is designed to select the most discriminative and relevant clip from each section and feed them into the classification module.
As shown in Fig.~\ref{fig:arch}, it consists of an {\bf observation network} ($F_s$) and a {\bf policy maker} that determine which clips to select.

The observation network aims to output the sampling probability of each clip.
Formally, it can be formulated as follows:
\begin{equation}
P^m = \phi (fc(F_s(c^m_1,\cdots,c^m_N))),
\label{equ:sampling}
\end{equation}
where $F_s$ denotes the backbone of observation network and $fc$ denotes a fully connected layer.
$F_s$ takes the clips $(c^m_1,\cdots,c^m_N)$ in section $S_m$ as inputs and produces the feature maps for each clip.
Then these feature maps are concatenated and fed into a fully connected layer $fc$ followed by softmax function $\phi$.
In sampling module, we can obtain the probabilities $P^m=\{p^m_1,...,p^m_N\}$ of sampling clips $C^m$ in section $S_m$, where $p^m_n\in[0,1]$.

After obtaining the probabilities $P^m$, the policy maker will determine which clip to keep.
Specifically, it will output an action $A^m=\{a^m_1, a^m_2, \cdots, a^m_N\}$ based on the estimated probabilities $P^m$. $A^m$ is a binary vector and $a^m_n \in \{0,1\}$, where $a^m_n = 1$ indicates the $n^{th}$ clip will be kept in section $S_m$, otherwise this clip will be discarded.
Since the sampling module selects one clip from each section, only one element in vector $A^m$ will be set to 1, and the rest will be set to 0.
During the training phase, we randomly select a clip from each section based on the probabilities $P^m$, and in the testing phase, we directly choose the clip with the highest probability.
This slight difference will enable the sampling module to explore more clips and add randomness to the training process to its generalization ability.

It should be noted that the policy makes the independent decision for each section. In this sense, we keep the same number of clips for each section and thus the selected clips can cover the entire video. In addition, this independence assumption will reduce the decision process in each section into a single-step Markov decision process. Therefore, there are only $N$ possible actions in each section, and the sampling module can also be viewed as a $N$-armed contextual bandit whose agent selects one from $N$ levers by observing the environment and gets a reward.
The rewards obtained by these selected $M$ clips will guide the training of the sampling module through the reinforcement learning algorithm (REINFORCE), which will be explained in section~\ref{sec:optimization}.

\subsection{Classification Module}
\label{sec:classification}

The classification module output the action prediction results of each selected clip.
Then rewards will be calculated from prediction and ground-truth label to guide the training of the sampling module.

3D Convolutional Neural Networks (3D CNNs)~\cite{TranBFTP15,R2+1D} have become an effective clip-level action recognition model.
Therefore, as shown in Fig.~\ref{fig:arch}, we choose 3D CNNs as the backbone of our classification module.
In experiments, to demonstrate the robustness of DSN framework, we try two kinds of network architectures: (1) more efficient one: R3D-18, and (2) more powerful one: R(2+1)D-34.
R3D is an inflated 3D ResNet whose input is a short clip of stacking multiple consecutive frames.
R(2+1)D is a competitive short-term classifier that decomposes a 3D convolution into separate spatial and temporal convolutions, making its optimization easier than original 3D CNN.
Different network architectures can demonstrate the generalization ability of DSN.

{\bf Prediction of DSN.} After the training of sampling module and classification module, our DSN keeps $M$ clips for each video. We first perform action recognition for each clip independently, and then report video-level results via an average fusion:
\begin{equation}
    \mathrm{DSN} (V) = \mathcal{H}(G^1,\cdots,G^M; \Phi) = \frac{1}{M}\sum_{m=1}^M \mathcal{F}(G^m;\Phi),
\label{avg}
\end{equation}
where $\mathcal{F}$ represents the 3D convolutional networks, $\Phi$ is its weight, and $G^m$ is the sampling module output.

\section{Learning of Dynamic Sampling Networks}
\label{sec:optimization}
In this section, we describe the details on the optimization of DSN framework.
First, we provide a detailed description on the training process of DSN framework.
Then, we present the important implementation details of learning DSN from trimmed and untrimmed videos.

\subsection{Optimization of DSN}

Since DSN framework is composed of the sampling module and the classification module coupled with each other, and involves a non-differentiable operation of selection, we need to design a special training scheme to optimize the two modules.
Specifically, we propose two strategies to optimize their model parameters. In the first scheme, we first train the classification module in advance and then tune the parameters of the sampling module by fixing the classification module parameters.
In the second scheme, we propose an alternate optimization scheme learning between classification module and sampling module.
In particular, as shown in Algorithm~\ref{alg:procedure}, if we fix the pre-trained classification module, we skip the process of updating the classification module parameters.
Otherwise, we first optimize the parameters $\Phi$ of the classification module by fixing the parameters of the sampling module.
Then, we update the model parameters $\Theta$ of the sampling module by using REINFORCE. The total iteration size is denoted by $epochs$.
We will give a detailed explanation of these two steps in the following.

{\bf Classification module optimization.} Given the training set $\{V_l, y^l\}_{l=1}^L$ where $y^l$ is the label of the video $V_l$.
The sampling module outputs $\mathbf{G}_l=\{G^1, G^2, \cdots, G^M\}$ for each video $V_l$ and the training objective classification module is defined as follow:
\begin{equation}
\ell(\Phi) = \sum_{l=1}^L \sum_{j=1}^J y_{j}^l \log H^l_j+ \lambda \| \Phi\|^2_2,
\end{equation}
where $H^l=\mathcal{H}(\mathbf{G}_l; \Phi)$ is the prediction score of video $V_l$ as defined in Equation (\ref{equ:h}) and Equation (\ref{avg}), $J$ is the total number of categories, $H^l_j$ is the $j^{th}$ element of $H^l$, $\lambda$ is a hyper-parameter controlling the cross-entropy loss and $L_2$ regularization term.
The above objective function can be easily optimized using stochastic gradient descent (SGD).
As mentioned in section~\ref{sec:classification}, if we sample more than one clip in $\mathbf{G}$, we will average these prediction scores of selected clips to obtain a video-level score, and use this score for training.

\floatname{algorithm}{Algorithm}
\renewcommand{\algorithmicrequire}{\textbf{Input:}}
\begin{algorithm}[t]
        \caption{The training algorithm of DSN framework.}
        \begin{algorithmic}[1]
        \Require Videos and their labels $\{V_l, y^l\}_{l=1}^L$
            \State Set $epochs$, $M$ and $N$
            \State Initialize the weight $\Theta$ and $\Phi$
            \For{ $s \leftarrow 1$ to $epochs$}
                \State Sample $(c_1^1,\dots,c_N^M)$ from $V_l$
                \If {\textbf{not} fixing the classification module}
                    \State $\mathbf{G}=(G_1,\dots,G_M) \leftarrow\mathcal{G}(c_1^1,\dots,c_N^M;\Theta)$
                    \State $H \leftarrow {\mathcal{H}(\mathbf{G};\Phi)}$
                    \State Update the parameters of $\mathcal{H}$, namely $\Phi$
                \EndIf
                \For{ $m \leftarrow 1$ to $M$}
                    \State $P^m = \phi (fc(F_s(c^m_1,...,c^m_N)))$
                    \State Obtain $A^m$ and $B^m$ based on $P^m$
                    \State Evaluate reward $R(A^m)$ and $R(B^m)$ by Eqn.~\ref{rein:rewards}
                    \State Backpropagate gradients computed by Eqn.~\ref{rein:final_expected_grad}
                \EndFor
            \EndFor
        \end{algorithmic}
\label{alg:procedure}
\end{algorithm}

{\bf Sampling module optimization.} After the training of the classification module, we move to optimize the parameter of the sampling module, namely, the weights of the observation network.
Since the selection operation is non-differentiable, we intend to use the reinforcement learning algorithm for optimization.
The objective of reinforcement learning is to learn a policy $\pi$ that decides actions by maximizing future rewards.
The reward is calculated by an evaluation metric that compares the generated prediction to the corresponding groundtruth.
Only the selected clips will be fed into the classification module.
To encourage the policy to be more discriminative, we associate the actions taken with the following reward function:
\begin{equation}
R(A^m)=
\left\{
\begin{array}{lr}
r^m   \quad \quad \text{if $G_m$ can be classified correctly,} \\
-\gamma \quad \text{ \qquad  otherwise.}
\end{array}
\right.
\label{rein:rewards}
\end{equation}
where $r^m$ represents the reward if the selected clip $G^m$ is classified correctly, which is the score of the correctly classified actions by the classification module.
On the other hand, we penalize it with a constant $-\gamma$ if the classification module cannot make a correct prediction for clip $G^m$.

The training objective of sampling module is to maximize the expected reward over all possible actions~\cite{SuttonB98}:
\begin{equation}
\begin{aligned}
J(\theta)=\mathbb{E}_{A^m\sim \pi_\theta}[R(A^m)].
\end{aligned}
\label{rein:expected_reward}
\end{equation}

\textbf{REINFORCE.} In order to maximize the expected reward of the sampling module, we use the Monte Carlo policy gradient algorithm REINFORCE~\cite{SuttonB98} to train the sampling module.
In contrast to value function methods, REINFORCE algorithm can select actions without consulting the value function.
With Monte-Carlo sampling using all samples in a mini-batch, the expected gradient can be approximated as follow:
\begin{equation}
\begin{aligned}
\nabla J(\theta)&=\mathbb{E}_{\pi} \left[R(A^m) \nabla_\theta\text{log} \pi(A^m|C^m,\theta)\right]\\
&=\mathbb{E}_{\pi} [R(A^m) \nabla_\theta \sum_{n=1}^N \text{log} (p^m_n a^m_n)].
\end{aligned}
\label{rein:expected_grad}
\end{equation}
where $\theta$ is the parameters of the sampling module and $C^m$ is the collection of sampled clips in section $S_m$.
As mentioned in section~\ref{sec:sampling}, to obtain $A^m$ in Equation (\ref{rein:expected_grad}), we sample the clip based on $P^m$.
When the clip $c_n^m$ is chosen, its corresponding $a_n^m$ will be set to $1$, and otherwise $0$.

\textbf{REINFORCE with Baseline.}
As REINFORCE may be of high variance and learn slowly, we utilize the REINFORCE with Baseline~\cite{SuttonB98} to reduce variance:
\begin{equation}
\begin{aligned}
\nabla J(\theta)&=\mathbb{E}_\pi \left[(R(A^m)-b) \nabla_\theta\text{log} \pi(A^m|C^m,\theta)\right], \\
\end{aligned}
\label{rein:expected_grad_bx}
\end{equation}
where the baseline $b$ can be an arbitrary function, as long as it does not depend on the action $A^m$.
The equation is still valid because we subtract the zero value.
It is proved as follows:
\begin{equation}
\begin{aligned}
&\mathbb{E}_\pi [ b \nabla_\theta\text{log} \pi(A^m|C^m,\theta)]\\
=&\int \pi(A^m|C^m,\theta)  b \nabla_\theta \text{log}\pi(A^m|C^m,\theta)d \pi\\
=&\int   b \nabla_\theta \pi(A^m|C^m,\theta)d \pi\\
=&b \nabla_\theta \int  \pi(A^m|C^m,\theta)d \pi\\
=&b \nabla_\theta 1=0.
\end{aligned}
\end{equation}

\textbf{Update Parameter.} We select the largest probability $p^*$ from $P^m$ to construct R($B^m$) as baseline. $B^m=\{b_1^m, ..., b_N^m\}$ is the action in which $b_n^m=1$ if $p_n^m=p^*$, and $b_n^m=0$ otherwise.
Using REINFORCE with the baseline, we rewrite the estimate of the gradient $\nabla J(\theta)$ in Equation(\ref{rein:expected_grad}) as:

\begin{equation}
\begin{aligned}
\nabla J(\theta)=&\mathbb{E}_\pi [ (R(A^m)-R(B^m)) \nabla_\theta \sum_{n=1}^N \text{log}(p^m_n a^m_n)].\\
\end{aligned}
\label{rein:final_expected_grad}
\end{equation}

To optimize the expected reward $J(\theta)$, we update weight $\theta$ by $\nabla J(\theta)$  as follow:
\begin{equation}
\begin{aligned}
\theta_{t+1}=\theta_t + \alpha \nabla J(\theta_t),
\end{aligned}
\label{rein:update_param}
\end{equation}
where $t$ is the iteration number and $\alpha$ is the learning rate.

{\bf Discussion on end-to-end training.}
DSN can be also optimized in an end-to-end scheme, that is, to update the parameters of two modules at the same time.
In this case, the classification module can hardly recognize the action correctly in the early stage of training. For any clip, the sampling module gets almost negative rewards and the whole system is hard to optimize properly.
Therefore, the alternating optimization between the two modules is a practical solution, and pre-training classification module before the optimization of the sampling module is helpful to improve the training efficiency of the sampling module in the early stages.

\begin{table*}[t]
\centering
\caption{{\bf Study on the DSN framework.} We list some competitive methods to demonstrate the effectiveness of DSN. The specific evaluation metric is described in section~\ref{subec:datasets}. For UCF101 and HMDB51, the results are reported on split1. $M:N$ is set as $1:3$ for training on trimmed datasets and $6:3$ on untrimmed datasets. The testing setting is the same with training.}
\begin{tabular}{c|c|c|c|c}
% \hline
\toprule[2pt]
Methond & UCF101 & HMDB51  & THUMOS14 & ActivityNet v1.3\\
\hline
Baseline (random) & 83.9\% & 53.6\% & 62.5\% & 62.0\%\\
Baseline (uniform) & 84.6\% & 56.8\% & 62.3\% & 62.6\%\\
Max response & 85.9\% & 55.9\% & 65.9\% & 65.1\%\\
\hline

DSN-f & 85.7\% & 57.8\% & 67.3\% & 64.9\%\\
DSN & \textbf{86.5}\% & \textbf{59.5}\% & \textbf{72.6}\% & \textbf{66.7}\%\\
\hline
Dense sampling & 88.6\% & 59.5\% & 65.6\% & 67.5\% \\
Oracle sampling & 90.1\% & 66.8\% & 73.1\% & 88.6\%\\
\toprule[2pt]
\end{tabular}
\label{tab:config_cmp}
\end{table*}

\subsection{DSN in Practice}
\label{sec:inference}
Finally, we give detailed descriptions about the implementations of DSN in practice. We use the notation of $ M: N $ to represent the sampling scheme in DSN that each video is divided into $ M $ sections of equal duration without overlap and $N$ clips are sampled from each section. Our DSN is a general framework, which could be applied on both trimmed and untrimmed video datasets. To fully investigate DSN framework, we report the results of DSN on both trimmed and untrimmed datasets in experiments.

Considering that the temporal duration of videos is different between trimmed and untrimmed datasets, we use different settings in our DSN implementation on the two types of datasets.
For {\bf trimmed video} datasets (e.g., UCF101, HMDB51), considering that the duration of each video is normally very short, we set $M=1$ and $N=3$ during the training phase. In testing, we use $M=4$ to yield better recognition results. The $N$ is consistent whether in training or testing.
For {\bf untrimmed video} datasets (e.g., THUMOS, ActivityNet), we try to use a larger $M$ under the constraint GPU memory. It is expected that action instances might only occupy a small portion of the whole untrimmed video. Larger numbers of sampling clips would help DSN to select the most action-relevant clips. Specifically,  when we utilize a lightweight backbone of ResNet3D-18,  we set $M=6$ and $N=3$ during training.
In testing, we keep $N=3$ and study with different values of $M$ to find a trade-off between performance and computational overhead.

\section{Experiments}
\label{sec:exp}

In this section, we first describe the datasets used in experiments and the implementation details.
Then we perform some exploration studies on different aspects of DSN.
After that, we compare the performance of DSN with state-of-the-art methods on four datasets.
Finally, we visualize the learned policy of DSN to provide some insights about our method.

\begin{figure*}[t]
  \centering
  \includegraphics[width=14cm]{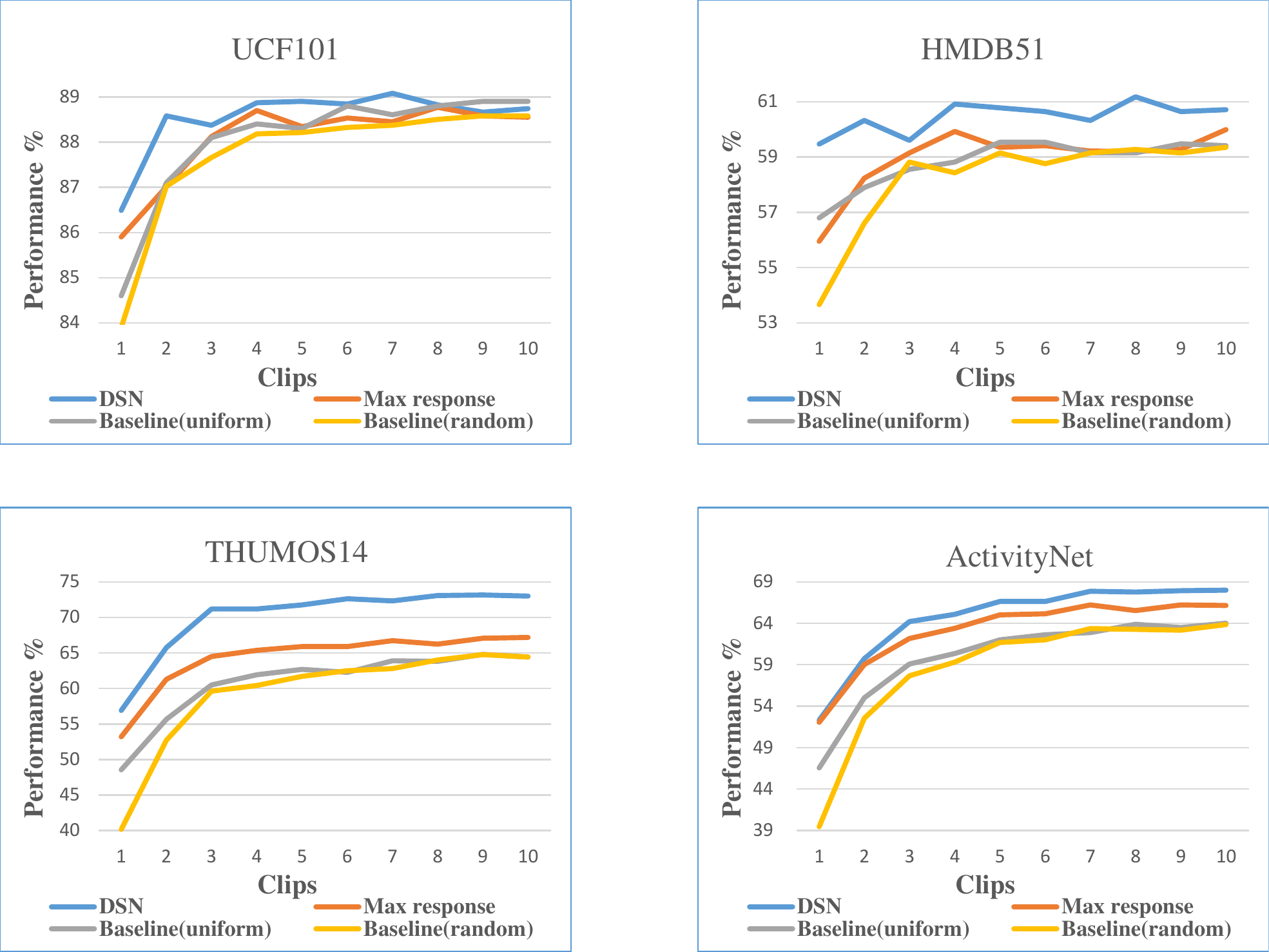}
  \caption{Results of different selection methods by using different numbers of clips. We report the Top-1 accuracy on the UCF101 (split1), HMDB51 (split1) and ActivityNet, and mAP on the THUMOS14 dataset. $M:N$ is set as $1:3$ for training on trimmed datasets and $6:3$ on untrimmed datasets. We set $N=3$ and vary $M$ from 1 to 10 for testing.}
  \label{fig:clip}
\end{figure*}

\subsection{Datasets}
\label{subec:datasets}

We report performance on two popular untrimmed datasets, \textbf{ActivityNet v1.3} ~\cite{anet} and \textbf{THUMOS14}~\cite{THUMOS14}.
ActivityNet v1.3 covers a wide range of complex human activities divided into 200 classes, which contains 10,024 training videos, 4,926 validation videos, and 5,044 testing videos.
Since annotations of action instances in untrimmed videos are hardly available in most cases, we only use video-level labels to train the model.
We adopt the Top-1 as the evaluation metric on ActivityNet v1.3 in ablation study and both Top-1 and mean average precision (mAP) as the evaluation metric comparing with the state-of-the-art models.
Due to the lack of annotated groundtruth of the test set, here we only report the performance on the validation dataset.
THUMOS14 has 101 classes for action recognition, which consists of four parts: training set, validation set, test set, and background set.
For the ablation study, we only train models on the validation set and evaluate their mAP on the test set. When comparing with other methods, we train DSN on the combination of the training and validation set.

We also validate DSN on two trimmed video datasets, \textbf{UCF101}~\cite{abs-1212-0402} and \textbf{HMDB51}~\cite{KuehneJGPS11}.
UCF101 has 101 action classes and 13K videos.
HMDB51 consists of 6.8K videos divided into 51 action categories. In trimmed video datasets, Top-1 is employed as an evaluation metric for all models.
We take the official evaluation protocol for these trimmed video datasets: adopting the three training/testing splits provided by the organizers and measuring average accuracy over these splits for evaluation.

%-------------------------------------------------------------------------

\subsection{Implementation Details}
For classification module, we use ResNet3D-18~\cite{abs-1708-05038} and R(2+1)D-34~\cite{R2+1D} as the backbone networks.
To make the sampling module as small as possible, we choose the ResNet2D-18~\cite{R2+1D} as the backbone of the observation network.
For ResNet2D-18 to process the same input length as R(2+1)D-34, we take the middle 8 frames in a single clip as input.
It is worth noting that the computation required for ResNet2D-18 is only 9.4\% of the ResNet3D-18 and 1.2\% of R(2+1)D-34, respectively.
In the experiment, we use SGD with momentum 0.9 as the optimizer.
ResNet3D-18, R(2+1)D-34 and ResNet2D-18 are pre-trained on Kinetics~\cite{CarreiraZ17}.
When training ResNet3D-18, the initial learning rate is set to 0.001, which is reduced at 45, 60 epochs by a factor of 10 and stops at 70 epochs.
Considering the R(2+1)D-34 is a heavier model, the whole training procedure is done in 120 $epochs$, starting at a learning rate of 0.001 and decreased at 90, 110 epochs by a factor of 10.
Video frames are scaled to the size of $128 \times171 $ and then patches are generated by randomly cropping window of size $ T \times 112\times112 $ with scale jittering.
The temporal dimension $T$ depends on the backbone architecture of classification module (i.e., 8 for ResNet3D-18 and 32 for R(2+1)D-34).
When training our models, the batch size is 128 for ResNet3D-18 and 32 for R(2+1)D-34.
We use Farneback’s method~\cite{optical-flow} to compute optical flow.

\subsection{Exploration Studies}
First, we are ready to perform ablation studies for DSN framework on trimmed and untrimmed video datasets.
In this exploration study, we choose ResNet3D-18 as the backbone network for the classification module and ResNet2D-18 for the observation network.
We use $M: N$ to denote training or testing setting that each video is divided into $M$ sections and each section has $N$ clips without overlap.
The sampling module keeps $M$ clips as inputs to the classification module.
All ablation experiments use RGB as input.

\begin{table}[t]
\centering
\caption{Exploration of the optimal sampling scheme in the training phase. All classification models use ResNet3D-18 as the backbone. The results are reported on split1.}
\begin{tabular}{ccccc}
\toprule[2pt]
&\multicolumn{2}{c}{UCF101} & \multicolumn{2}{c}{HMDB51} \\
\hline
Training Setting &Top-1&Top-5&Top-1&Top-5\\
\hline
1:3& 88.4\% & 98.5\% & 59.6\% & 88.0\%\\
\hline
1:6 & 88.7\% & 98.8\% & 59.2\% & 88.6\%\\
\hline
3:3 & 89.1\% & 98.6\% & 61.5\% & 89.8\%\\
\bottomrule[2pt]
\end{tabular}
\label{tab:train}
\end{table}

\begin{table}[h]
\centering
\caption{Comparison of FLOPS and accuracy on HMDB51. The accuracy is reported on split1.}
\begin{tabular}{ccccc}
\toprule[2pt]
Model & Clips & FLOPs & Top-1\\
\midrule
R3D &10& 203.6G &  59.5\% \\
R3D \& DSN ($M=1$)&4& 104.6G &60.9\%\\
R3D \& DSN ($M=3$)&3& 78.5G&61.5\% \\
\midrule
R(2+1)D &10& 1528.8G & 75.5\% \\
R(2+1)D \& DSN ($M=1$)&4& 631.2G& 75.8\%\\
\bottomrule[2pt]
\end{tabular}
\vspace{2mm}
\label{tab:flops_hmdb}
\end{table}

\begin{table}[h]
\centering
\caption{Comparison of FLOPS and accuracy on THUMOS14. The models are trained on the training set and validation set, and tested on the test set.}
\begin{tabular}{ccccc}
\toprule[2pt]
Model & Clips & FLOPs &  mAP\\
\midrule
R3D \& TSN ($S=3$) &20& 407.2G & 73.6\% \\
R3D \& DSN ($M=3$)&6& 157.0G& 75.2\%\\
\midrule
R(2+1)D &20& 3057.6G & 81.9\%\\
R(2+1)D \& DSN ($M=1$)&6& 946.8G& 83.1\%\\
\bottomrule[2pt]
\end{tabular}
\vspace{2mm}
\label{tab:flops_thumos}
\end{table}

\begin{table*}[t]
\centering
\caption{Comparisons with the current state-of-the-art models on THUMOS14. The DSN models are trained on the training set and validation set and tested on the test set.}
\begin{tabular}{l|c|c|c}
\toprule[2pt]
Method &Input &Clips& mAP \\
\hline
RGB+EMV~\cite{ZhangWW0W16}&RGB+FLOW&-& 61.5\%\\
Two Stream~\cite{SimonyanZ14}&RGB+FLOW&25& 66.1\%\\
iDT+CNN~\cite{wang2014action}&RGB+iDT&ALL&62.0\%\\
%Objects~\cite{JainGS15} & 44.7\% \\
Motion~\cite{JainGS15} &HOG+HOF+MBH&ALL& 63.1\% \\
Objects+Motion~\cite{JainGS15} &RGB+HOG+HOF+MBH&ALL& 71.6\% \\
Jain \emph{et al.}~\cite{jain2014university}&HOG+HOF+MBH&ALL& 71.0\% \\
ELM~\cite{varol2014extreme} &HOG+HOF+MBH&ALL& 62.3\%\\
%TSN (3 seg)~\cite{TSN-J}& 78.8\%\\
TSN~\cite{TSN-J}&RGB+FLOW&ALL& 80.1\%\\
UntrimmedNet (hard)~\cite{WangXLG17}  &RGB+FLOW&ALL& 81.2\%\\
UntrimmedNet (soft)~\cite{WangXLG17} &RGB+FLOW&ALL&\textbf{82.2\%}\\
\hline
R3D-RGB \& TSN (6 seg)~ &RGB& 20 & 73.6\%\\
\textbf{R3D-RGB \& DSN} ($M=6$)~ &RGB& 6&\textbf{75.2\%} \\
\hline
R(2+1)D&RGB&20& 81.9\%\\
R(2+1)D&FLOW&20& 73.5\%\\
R(2+1)D&RGB+FLOW&20& 82.6\%\\
% \hline
R(2+1)D \& DSN ($M=1$)~ &RGB&6& 83.1\% \\
R(2+1)D \& DSN ($M=1$)~ &FLOW&6& 78.4\% \\
\textbf{R(2+1)D \& DSN} ($M=1$)~ &RGB+FLOW&6& \textbf{84.7\%} \\

\bottomrule[2pt]
\end{tabular}
\label{tab:state_of_the_art_thumos}
\end{table*}

\begin{table*}[t]
\centering
\caption{Comparisons with the state-of-the-art models on ActivityNet v1.3. We list some competitive models to compare with our DSN. The $M$ in the first column denotes how much clips are used during the DSN training.}
\begin{tabular}{l|c|c|c|c|c}
\toprule[2pt]
Method &Backbone & Pre-trained& Clips&Top-1 & mAP \\
\hline
IDT~\cite{WangS13a}&-&ImageNet&ALL& 64.7\% & 68.7\%\\
C3D~\cite{QiuYM17}&-&Sports1M&10& 65.8\% & 67.7\%\\
P3D~\cite{QiuYM17}&ResNet-152&ImageNet&20& 75.1\% & 78.9\%\\
RRA~\cite{ZhuTZLYDM18}&ResNet-152&ImageNet&25& 78.8\% & 83.4\%\\
MARL~\cite{abs-1907-13369}&ResNet-152&ImageNet&25& 79.8\% & 83.8\%\\
MARL~\cite{abs-1907-13369}&ResNet-152&Kinetics&25& 80.2\% & 83.5\%\\
MARL~\cite{abs-1907-13369}&SEResNeXt152&Kinetics&25& \textbf{85.7\%} & \textbf{90.1\%}\\
\hline
R3D-RGB \& TSN (6 seg)&\multirow{3}*{ResNet-18}&\multirow{3}*{Kinetics}&20&65.6\% & 68.7\%\\
R3D-RGB \& DSN ($M=6$)&~&~&6&66.7\%&71.4\%\\
\textbf{R3D-RGB \& DSN} ($M=6$)&~&~&10&\textbf{68.0\%}&\textbf{71.7\%}\\
\hline
R(2+1)D-RGB&\multirow{9}*{ResNet-34}&\multirow{9}*{Kinetics}&20&78.2\% & 83.6\%\\
R(2+1)D-FLOW&~&~&20&76.9\%&80.7\%\\
R(2+1)D-RGB+FLOW&~&~&20&81.3\%&87.1\%\\
R(2+1)D-RGB \& DSN ($M=1$)&~&~&6&76.5\%&82.6\%\\
R(2+1)D-FLOW \& DSN ($M=1$)&~&~&6&78.2\%&82.9\%\\
R(2+1)D-RGB+FLOW \& DSN ($M=1$)&~&~&6&82.1\%&87.8\%\\
R(2+1)D-RGB \& DSN ($M=1$)&~&~&10&78.5\%&83.5\%\\
R(2+1)D-FLOW \& DSN ($M=1$)&~&~&10&79.1\%&84.3\%\\
\textbf{R(2+1)D-RGB+FLOW \& DSN} ($M=1$)&~&~&10&\textbf{82.6\%}&\textbf{87.8\%}\\
\bottomrule[2pt]
\end{tabular}
\label{tab:state_of_the_art_act}
\end{table*}

\begin{table*}[t]
\centering
\caption{Comparison with the current state-of-the-art models on UCF101 and HMDB51. The models equipped with DSN are trained by sampling scheme $M:3$ where $M$ is presented in the table.
The accuracy is reported as average over the three splits. We only compare with the models using RGB as input. }
\begin{tabular}{l|c|c|c|c|c}
\toprule[2pt]
Method & Input & Clips & Backbone & UCF101 & HMDB51 \\
\hline
TSN RGB~\cite{TSN-J}& $[1\times 3 \times 224 \times 224] $&25&Inception V2&91.1\%&-\\
TSN RGB~\cite{TSN-J}& $[1\times 3 \times 229 \times 229] $&25&Inception V3&93.2\%&-\\
\hline
C3D~\cite{TranBFTP15}&$[16 \times 3 \times 112 \times 112]$&10&VGGNet-11&82.3\%&51.6\%\\
C3D~\cite{abs-1708-05038}&$[16 \times 3 \times 112 \times 112]$&10&ResNet-18&85.8\%&54.9\%\\
\hline
I3D-RGB~\cite{CarreiraZ17}& $[64 \times 3 \times 224 \times 224] $&ALL&Inception V1&95.6\%&74.8\%\\
\hline
P3D~\cite{QiuYM17}&$[32 \times 3 \times 299 \times 299]$&20&ResNet-152&88.6\%&-\\
\hline
MF-Net~\cite{ChenKLYF18}& $[16 \times 3 \times 224 \times 224]$&50&-&96.0\%&74.6\%\\
\hline
ARTNet with TSN~\cite{ArtNet}& $[24 \times 3 \times 112 \times 112]$&25&ResNet-18 &94.3\%&70.9\%\\
\hline
S3D-G~\cite{s3d}& $[64 \times 3 \times 224 \times 224]$&ALL&Inception &\textbf{96.8\%}&\textbf{75.9\%}\\
\hline
R3D-RGB & $[8 \times 3 \times 112 \times 112] $ & 10&\multirow{3}*{ResNet-18} & 88.6\% & 58.8\%\\
R3D-RGB \& DSN ($M=1$)& $[8 \times 3 \times 112 \times 112] $ &4&~& 88.7\% & 59.7\%\\
R3D-RGB \& DSN ($M=3$)& $[8 \times 3 \times 112 \times 112] $ &3&~& 89.5\% & 61.6\%\\
\hline
R(2+1)D-RGB~\cite{R2+1D}& $[32 \times 3 \times 112 \times 112] $ &10&\multirow{3}*{ResNet-34}& 96.8\% & 74.5\%\\
R(2+1)D-RGB (\textbf{ours})& $[32 \times 3 \times 112 \times 112] $ &10&~& 96.7\% & 74.8\%\\
R(2+1)D-RGB \& DSN ($M=1$) & $[32 \times 3 \times 112 \times 112] $  &4&~& \textbf{96.8}\% & \textbf{75.5}\%\\
\bottomrule[2pt]
\end{tabular}
\vspace{2mm}
\label{tab:state_of_the_art}
\end{table*}

\textbf{Study on DSN.}
To demonstrate the effectiveness of DSN framework, we compare it to five different methods and another different training method of DSN mentioned in section~\ref{sec:optimization} called fixed DSN.
In this experiment, DSN is trained with the setting of $1:3$ for trimmed video and $6:3$ for untrimmed video respectively, and to be consistent with training, we test with the same setting.

We choose random sampling and uniform sampling as the baseline.
They use the same clip-level classifier, randomly sample $M$ clips and uniformly sample $M$ clips from the whole video in the test phase, respectively.
We also compare DSN with a competitive method called max response sampling~\cite{HuangRMTPFN18} which trains another observation network and uses the maximum classification score to sample clips.
Then we introduce the fixed DSN in section~\ref{sec:optimization} which trains the sampling module with a trained classification module and denote it as DSN-f.
Finally, we list the results of dense sampling and oracle sampling~\cite{HuangRMTPFN18} which are the most intuitive measurement of the performance of the classifier.
All results are shown in Table~\ref{tab:config_cmp}.

The clip-level classifiers of the baseline methods are trained by randomly sampled frames.
When $M>1$, we average the predictions of all clips to get the final predictions and train the baseline.
As shown in Table~\ref{tab:config_cmp}, on UCF101 and HMDB51, DSN outperforms the random baseline  by 2.6\%  and 5.9\% and outperforms the uniform baseline by 1.9\% and 2.7\%, respectively.
For untrimmed video tests, DSN performance improvements are more evident.
On THUMOS14 and ActivityNet v1.3, DSN is 10.1\% and 4.7\% higher than the random baseline and 10.3\% and 4.1\% higher than the uniform baseline.
The promising results demonstrate that different sampling policies have a huge impact on the classification performance, and also validate that the sampling module has indeed learned the effective sampling policies.

The method called max response sampling trains another observation network independently and outputs the classification score for each clip.
The max response is defined as the maximum classification score of all the classes.
We only select the clip with the highest max response in each section as input to the classification module during testing.
For a fair comparison, we use ResNet2D-18 as the backbone of the observation network and sample $M$ clips from $M \times N$ according to maximum score during testing.
We observe that DSN is better than max response sampling by 0.6\% on UCF101, 3.6\% on HMDB51, 6.7\% on THUMOS14 and 1.6\% on ActivityNet v1.3.
This demonstrates that simply using max response as the sampling policy obtains a very limited performance improvement.

DSN-f is a simplified training version of DSN.
Compared to DSN, the parameters of the classification module of DSN-f are fixed and no longer updated during training.
The parameters of the classification module of DSN-f are consistent with the baseline.
Without joint training, the performance of DSN-f is lower than that of DSN.
On the four datasets, DSN-f is decreased by 0.8\%, 1.7\%, 5.3\%, and 1.8\%, respectively.
We analyze that the joint training of two modules alternately enables the sampling module and the classification module to perform their duties, focus more on the learning of their respective tasks, and at the same time better couple with each other.
Therefore, compared with DSN-f, training DSN model by joint training is more adequate.
This fully demonstrates that joint training is indispensable.

The dense sampling results are the experimental upper bound of the classifier in practice.
Currently, the state-of-the-art results of most models are obtained by dense sampling.
For the trimmed video and untrimmed video datasets, we take 10 clips and 20 clips respectively as input to the classification module.
On the trimmed video dataset, DSN uses only one clip as input and achieves performance comparable to dense sampling using 10 clips.
On the untrimmed video dataset, DSN using 6 clips as input not only achieves performance comparable to the dense sampling using 20 clips on ActivityNet v1.3 but also has higher mAP on THUMOS14 than the dense sampling.
Oracle sampling is the theoretical upper bound of the sampling policy.
As long as there is a clip that can be correctly classified by the classifier, oracle sampling will select this clip as the input to the classifier.
The results are almost impossible to achieve in the experiment.
There is still a gap between DSN and oracle sampling, which shows that there is room for further improvement and optimization of the clip sampling policy.

\textbf{Study on different numbers of sampled clips.}
We have demonstrated the effectiveness of DSN.
Now, we are ready to explore the performance curve of DSN by varying the number of sampled clips.
We gradually increase the number of sampled clips from 1 to 10 and compare it to the baseline and max response sampling mentioned in Table~\ref{tab:config_cmp}.
The experimental results are shown in Fig.~\ref{fig:clip}.

For trimmed video, when the number of clips is greater than 4, the advantages of DSN over other methods will gradually decrease.
We analyze that due to most of the useless frames of the trimmed video have been trimmed and the duration of trimmed videos is short in general, as the number of clips increases,  most of the frames will be selected, and all methods will gradually degenerate into dense sampling.
For untrimmed video datasets, DSN has an advantage over other methods when using a small number of clips.
But when the number of clips is greater than 3 and 5 on THUMOS14 and ActivityNet, the performance gain from increasing clip number gradually decreases.
The reason is that the useful frames of untrimmed videos are distributed unevenly and sparsely, and it is difficult to cover them by simply increasing the number of clips.
Experimental results show that DSN can almost reach its peak performance when computing a small number of clips, so in practical deployment, when the computing resources are limited, computing a small number of clips can achieve very good results.

\begin{figure*}[t]
  \centering
  \includegraphics[width=1.0\textwidth]{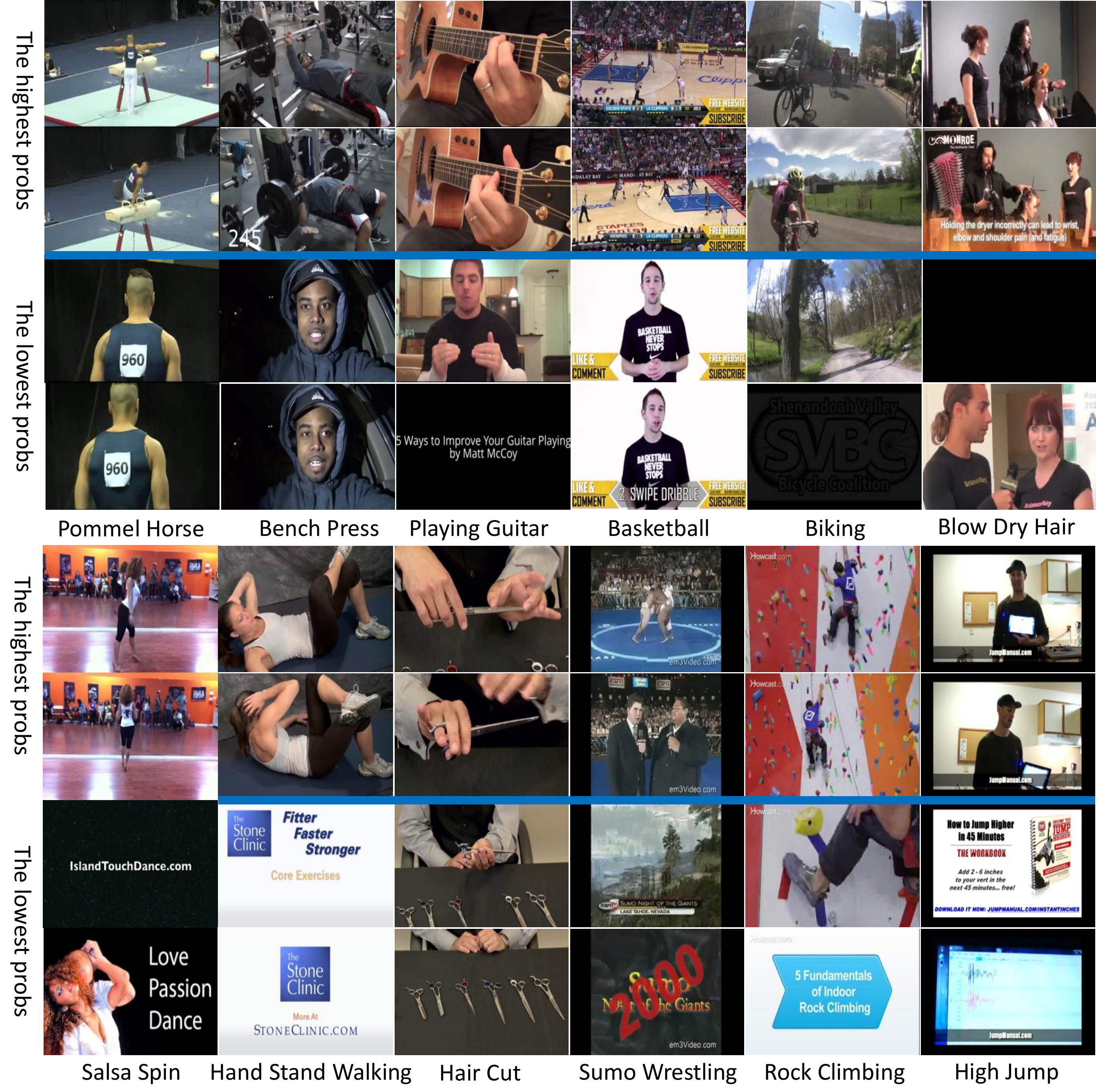}
  \caption{Visualization of the frames chosen by the sampling module on THUMOS14. Each column represents frames sampled from the same video. For each video, we show two frames with the highest confidence on the top and two frames with the lowest confidence on the bottom, respectively.}
  \label{fig:visualization}
  \vspace{-2mm}
\end{figure*}

\textbf{Study on training setting.}
After evaluating the performance trend of DSN, we are ready to investigate the effect of the different number of sections and the number of clips in each section in the training sampling scheme.
In training, we use the training sampling schemes of $1:3$, $1:6$, and $3:3$. In testing, the $N$ is the same as training and $M$ is set to 3 for fair comparison between different sampling schemes.
The results are shown in Table~\ref{tab:train}.
When $M=1$, simply increasing the number of clips in each section does not bring about significant performance improvement.
The reason is that increasing the number of segments means enlarging the exploration space of reinforcement learning, which may lead to under-fitting, so sampling module needs more epoch to explore.
With the same number of training epochs, the sampling module is difficult to get enough training.
When we increased $M$ from 1 to 3, we achieved 0.7\% and 1.9\% performance improvement on UCF101 and HMDB51 respectively.
This is because more sections mean that the sampling module and the classification module can obtain more global semantic information.
This avoids the imbalance of the learning samples caused by repeated local sampling during the training phase.
At the same time, in the training phase of the classification module, based on the conclusion of TSN~\cite{TSN-J}, the effect of the multi-segments fusion training is significantly better than the single-segment training.
Therefore, when computational resources are sufficient, increasing the number of sections can obtain better performance.

\textbf{Quantitative analysis in FLOPs.}
We perform experiments on computational overhead and the performance of DSN.
In the experiments, R3D uses ResNet3D-18 as the backbone and R(2+1)D use ResNet3D-34 as the backbone, respectively.
As shown in Table~\ref{tab:flops_hmdb}, on HMDB51, R3D with DSN uses only 3 clips and one-third of the baseline FLOPS, and the accuracy is 2\% higher than the baseline.
R(2+1)D uses only 4 clips and less than half of the baseline FLOPs and achieves performance comparable to baseline.
The effect is more obvious on THUMOS14.
As shown in Table~\ref{tab:flops_thumos}, R3D with DSN uses only 6 clips, which is 1.6\% higher than the baseline using 20 clips, while saving more than half of FLOPs.
R(2+1)D requires only one-third of FLOPs and achieves accuracy improvement of 1.2\%.
This indicates that DSN framework has great advantages in improving accuracy and saving computational cost.

\subsection{Comparison with the State of the Art}
We compare DSN with the current state-of-the-art methods.
In the experiments, R3D uses ResNet3D-18 as the backbone and R(2+1)D uses ResNet3D-34 as the backbone, respectively.
The number of clips used by other methods is also given in the table.

\textbf{Results on untrimmed video datasets.}
We list the performance of DSN models and other state-of-the-art approaches on THUMOS14 and ActivityNet v1.3.
The results of both RGB and optical flow are shown in Table~\ref{tab:state_of_the_art_thumos} and Table~\ref{tab:state_of_the_art_act}.
DSN is trained with the setting of $6:3$ for R3D and $1:3$ for R(2+1)D.
R3D and R(2+1)D use the same sampling scheme $N=3$ and $M=6$ or $M=10$ for testing.

As shown in Table~\ref{tab:state_of_the_art_thumos}, on THUMOS14, for R3D and R(2+1)D, DSN only uses 6 clips, and their mAP is 1.6\% and 1.2\% higher than the dense sampling when using RGB as input.
The optical flow and two-stream results of R(2+1)D are higher than dense sampling by 4.9\% and 2.1\%, respectively.
R(2+1)D trained with DSN only uses RGB and 6 clips as input, and it achieves better performance than other methods of two-stream results.
We give more experimental results on ActivityNet v1.3 in Table~\ref{tab:state_of_the_art_act}.
For R3D, both DSN and baseline use 6 clips for training。
When using RGB as input and sampling 6 clips from each video, the Top-1 accuracy of DSN is 1.1\% higher than the baseline using 20 clips.
When only half of the input of the baseline is used, that is, 10 clips, the Top-1 accuracy is directly increased by 2.4\%.
For R(2+1)D, both DSN and baseline use only one clip for training.
The Top-1 accuracy of DSN using only 10 clips is 0.3\% higher than the baseline using 20 clips.
It has also achieved very obvious effects on the optical flow that outperform dense sampling by 1.3\% and 2.2\% by using 6 clips and 10 clips, respectively. These experimental results demonstrate that DSN framework only uses video-level supervision information for training, which allows the sampling module to select useful clips in the video.

\textbf{Results on trimmed video datasets.}
We furthermore compare the performance of our method with the state-of-the-art approaches on UCF101 and HMDB51 in Table~\ref{tab:state_of_the_art}.
We only list the accuracy of the models using RGB as input.
The DSN models are trained with $N=3$, and testing sampling scheme is set as $4:3$.

On UCF101, for R3D backbone with $1:3$ training scheme, the Top-1 accuracy of DSN using 4 clips is comparable to the baseline using 10 clips.
When using $3:3$ training scheme, the input of 3 clips can improve accuracy by nearly 1\%.
For R(2+1)D, DSN uses only 4 clips to get the same performance as the baseline using 10 clips.
The experimental results on HMDB51 are more obvious.
After R3D and R(2+1)D are trained by DSN and with $1:3$ training scheme, the results of the model using 4 clips are 0.9\% and 0.7\% higher respectively than the baseline using 10 clips. These experimental results demonstrate that although the trimmed datasets have removed most of the irrelevant frames that do not contain actions, DSN framework can still filter out the useless clips so that the performance is further improved when the performance has already saturated.

\subsection{Visualization and Discussion}
After analyzing the quantitative results in both aspects of recognition accuracy and computational overhead, we present the visualization results of the sampling policies in this section.
Fig.~\ref{fig:visualization} illustrates sampling frames sorted by the output of the observation network in the sampling module.
We show the top 2 frames with the highest confidence and last 2 frames with the lowest confidence.
It is seen that the sampling module in DSN framework can select discriminative action clips and avoid irrelevant clips that might degrade the classifier training.
For some videos, the duration of action is usually a short clip through the whole video but interspersed with some irrelevant pictures.
For example, sometimes it needs some useless preparations before the action happens.
Even during the action is happening, the camera shot may turn to other irrelevant scenes.
Those useless clips not only bring a large waste of computing but also degenerate the performance in final recognition.
From Fig.~\ref{fig:visualization}, this observation shows that the motivation of DSN is of practical significance that it is necessary to sample frames based on some optimal policy to train a better clip-level action classifier.

\section{Conclusions and Future Work}
\label{sec:con}
In this paper, we have presented a new framework for learning action recognition models, called {\em Dynamic Sampling Networks} (DSN).
DSN is composed of a sampling module and a classification module, whose objectives are to dynamically select discriminative clips given an input video and to perform action recognition on the selected clips, respectively.
The sampling module in DSN could be optimized within an associative reinforcement learning algorithm.
The classification module can be optimized by SGD with cross-entropy loss.
Meanwhile, we provide two different schemes to optimize DSN framework.
One iteratively optimizes the sampling module and classification module, while the other only trains the sampling module while fixing the classification module.
We conduct extensive experiments on several action recognition datasets, including trimmed and untrimmed video recognition.
As demonstrated on four challenging action recognition benchmarks, DSN can greatly reduce the computational overhead yet is still able to achieve slightly better or comparable accuracy to the previous state-of-the-art approaches.

In the future, we still have many directions to improve the DSN framework.
Compared with the current alternating optimization scheme between sampling module and classification module, we can utilize the reparameterization trick and Gumbel-Softmax to discretize the output of the sampling module, so as to solve the non-differentiable problem of clip selection operation and realize an end-to-end training of the DSN framework.
In addition, the lastest reinforcement learning algorithms such as PPO~\cite{ppo} could be used to improve sample efficiency and robustness.
Finally, we can also use temporal models such as LSTM~\cite{lstm} to better exploit temporal information across different sections for better clip selection.

% or


% \section*{Acknowledgment}


\ifCLASSOPTIONcaptionsoff
  \newpage
\fi



%\bibliographystyle{IEEEtran}

\bibliographystyle{IEEEtran}
\bibliography{egbib}


% \end{IEEEbiography}


%\newpage



%\vfill


\end{document}